# Event-based Heterogeneous Information Processing for Online Vision-based Obstacle Detection and Localization

Reza Ahmadvand

*School of Electrical and Computer Engineering, University of Oklahoma, Norman, OK 73071, US, Email: iamrezaahmadvand1@ou.edu*

Sarah Safura Sharif

*School of Electrical and Computer Engineering, University of Oklahoma, Norman, OK 73071, US, Email: s.sh@ou.edu*

Yaser Mike Banad

*School of Electrical and Computer Engineering, University of Oklahoma, Norman, OK 73071, US, Email: bana@ou.edu*

This paper introduces a novel framework for robotic vision-based navigation that integrates Hybrid Neural Networks (HNNs) with Spiking Neural Network (SNN)-based filtering to enhance situational awareness for unmodeled obstacle detection and localization. By leveraging the complementary strengths of Artificial Neural Networks (ANNs) and SNNs, the system achieves both accurate environmental understanding and fast, energy-efficient processing. The proposed architecture employs a dual-pathway approach: an ANN component processes static spatial features at low frequency, while an SNN component handles dynamic, event-based sensor data in real time. Unlike conventional hybrid architectures that rely on domain conversion mechanisms, our system incorporates a pre-developed SNN-based filter that directly utilizes spike-encoded inputs for localization and state estimation. Detected anomalies are validated using contextual information from the ANN pathway and continuously tracked to support anticipatory navigation strategies. Simulation results demonstrate that the proposed method offers acceptable detection accuracy while maintaining computational efficiency close to SNN-only implementations, which operate at a fraction of the resource cost. This framework represents a significant advancement in neuromorphic navigation systems for robots operating in unpredictable and dynamic environments.

## I. Introduction

Autonomous robotic systems deployed in complex and dynamic environments must possess robust and energy-efficient situational awareness capabilities. A fundamental requirement in such systems is the ability to detect, localize, and track unmodeled intruders in real time, often under limited computational budgets and uncertain sensor conditions. Traditional Artificial Neural Networks (ANNs) have shown effectiveness in spatial feature extraction but suffer from synchronous, frame-based processing that incurs high latency and energy demands, making them suboptimal for real-time edge deployment [1,2]. On the other hand, Spiking Neural Networks (SNNs) offer event-driven computation and low power usage, making them highly suitable for dynamic sensory processing [3]. However, their ability to represent static spatial context remains limited [4]. To address these limitations, Hybrid Neural Networks (HNNs) have emerged as a promising solution by combining the complementary strengths of ANNs and SNNs. The Hybrid Sensing Network (HSN) framework introduced in [5,6] adopts a dual-pathway design, where an ANN handles low-frequency frame-based data while a recurrent SNN processes high-frequency spike-encoded input from event-based sensors like Dynamic Vision Cameras (DVCs). These two pathways are fused via a Hybrid Unit (HU), enabling synchronized spatiotemporal feature integration [5]. This structure has shown promising results in detection tasks under changing visual dynamics. Despite this architectural efficiency, reliable state estimation in the presence of uncertainty remains a challenge. To this end, we integrate a neuromorphic estimator—SNN-EMSIF—recently proposed in [7], which combines the Extended Modified Sliding Innovation Filter (EMSIF) [6] with the event-based, low-power nature of SNNs. This framework embeds system dynamics directly into the weight matrices of Leaky Integrate-and-Fire (LIF) neurons, avoiding the need for online learning. Additionally, the SNN-EMSIF design ensures robustness to modeling errors and partial observability [8] and remains resilient to neuron silencing due to its distributed and parallel structure. In this paper, we propose a unified hybrid neural framework that fuses the perception capabilities of HSN with the robustness of SNN-EMSIF filtering. The resulting system performs intruder detection and localization with high accuracy, low latency, and minimal computational burden—making it suitable for deployment in real-time robotics in dynamically changing environments in the presence of unmodeled moving intruders.

## II. Theory

This section presents the deployment architecture for the implementation of the proposed method, then outlines some theoretical preliminaries related to the utilized methods. Here, we present a novel framework for robot vision-based navigation systems that integrates Hybrid Neural Networks (HNNs) with Spiking Neural Network-based Filtering for enhanced situational awareness and unmodeled obstacle detection. Figure 1 schematically depicts the considered architecture and dual data pathways.

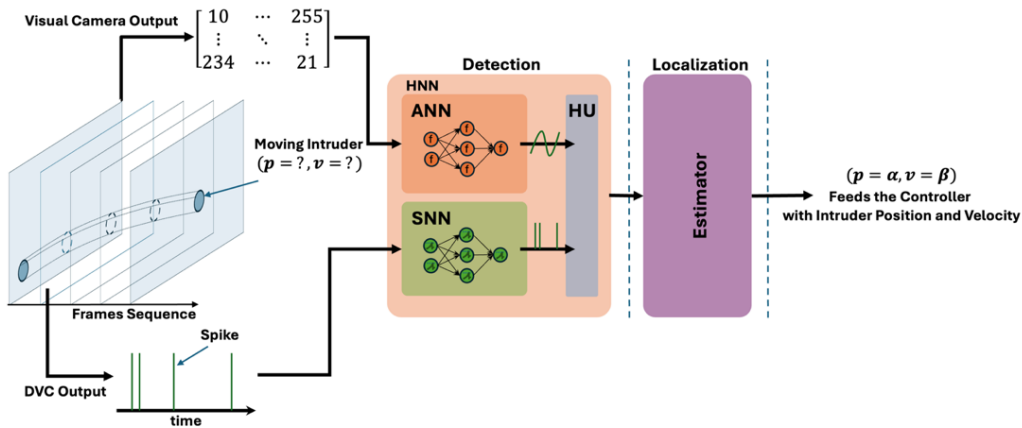

**Fig. 1 Hybrid Neural Framework for real-time intruder detection and localization, combining visual frames and DVC spike events processed by ANN and SNN components. Outputs are fused and used for state estimation and navigation**

Our approach combines the complementary strengths of Artificial Neural Networks (ANNs) in accurate stationary data processing and Spiking Neural Networks (SNNs) in low-latency real-time dynamic data processing. The proposed architecture features a dual-pathway processing system: an ANN component that handles static environmental features and spatial relationships, paired with an SNN component that processes temporal dynamics and event-based information. Rather than using hybrid units for domain conversion between the real-time data flow from the SNNs to the filter we leverage pre-developed SNN-based filter that naturally process spike-based information for the localization purposes. Figure 1 shows our implementation architecture. Our system processes sensor data through parallel computational pathways: high-resolution camera images of environment flow through the ANN path, generating comprehensive feature maps of the static environment at a low frequency, while the dynamic vision camera (DVC) which is an event-based sensor feeds the SNN path, producing real-time spike-encoded representations at substantially higher frequencies. The information flow maintains an efficient division between static features (f) processed by the ANN component and dynamic changes (Δf) handled by the SNN component. The detection pipeline begins with high frequency pre-attentive SNN processing that monitors for temporal anomalies in DVC data, triggering focused computational attention when potential obstacles are detected. These detected anomalies undergo spatiotemporal integration through the SNN-based filter, which maintains state estimates for the localization purposes. Simultaneously, contextual information from the ANN pathway validates whether detections represent truly unmodeled objects or unexpected environmental variations. Confirmed obstacles are continuously tracked with refined state estimates, enabling anticipatory navigation decisions based on predicted trajectories for obstacles. Here, The ANN path is responsible for extracting static spatial features from conventional visual frames. It operates based on classical feedforward dynamics, where each layer's output is computed as:

$$a_n = f_a\left(\sum_i w_i a_{n-1}\right) \quad (1)$$

Here, $a_n$ represents the activation at layer $n$, and $w_i$ stands for the learnable weights, and $f_a(.)$ denotes a nonlinear activation function. This formulation enables high-precision encoding of frame-level object characteristics. On the other hand, to encounter with dynamic changes in the environment, such as the motion of an unmodeled intruder, a SNN governed by LIF dynamics has been implemented:

$$\tau_u \frac{du_n}{dt} = -g(u_n(t)) + \sum_i w_i s_{n-1}^i(t) \quad (2)$$

$$S_n(t) = H(u_n^t - v_{th}) \quad (3)$$

In the above, $u_n(t)$ denotes the membrane potential of $n^{th}$ neuron at time $t$, and $g(.)$ represents the leakage function, and $S_n(t)$ is the spike output defined by the Heaviside step function $H(.)$. When the membrane potential exceeds a threshold $v_{th}$ a spike is emitted by the neuron.

### III. Simulation Result

Preliminary evaluations in simulation environments demonstrate our framework achieves good results while it has significant advantages in low latency, computational efficiency, and accuracy in the complex scenarios consisting of more than one obstacle. Here, to evaluate the proposed framework it has been applied into the dataset released by [5]. Figure 2 shows three different frames of a rotating disk video with three objects on it which has been used as our dataset for object detection.

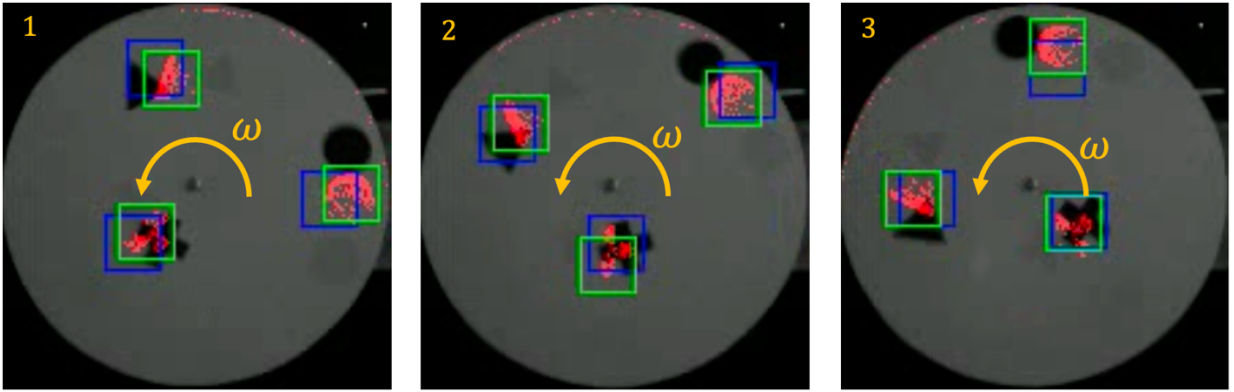

**Fig. 2 different three different frames of the implemented dataset for the rotating disk with three different objects.**

Figure 3 demonstrated the obtained results for the implementation of our proposed framework on the above-mentioned dataset. Here, Figure 3(a) compares the obtained estimated trajectories form the SNN-based filter integrated to the HNN (solid lines) with their true trajectories (dashed lines) and it is apparently visible that at the first steps, it is experiencing errors for the position estimation and also it starts with a wrong estimation on the velocity vectors (black vectors) for all the objects. Figure 3(b) shows the estimation of the angular velocity which shows an acceptable accuracy of the estimation with some fluctuations around true angular velocity which is equal to 0.05 rad/frame. Figure 3(c) shows the estimation error history for the estimated trajectory with respect to the true trajectories of the objects. Note that, the objects 1, 2 and 3 are cross, triangle and circle respectively in the original implemented dataset.

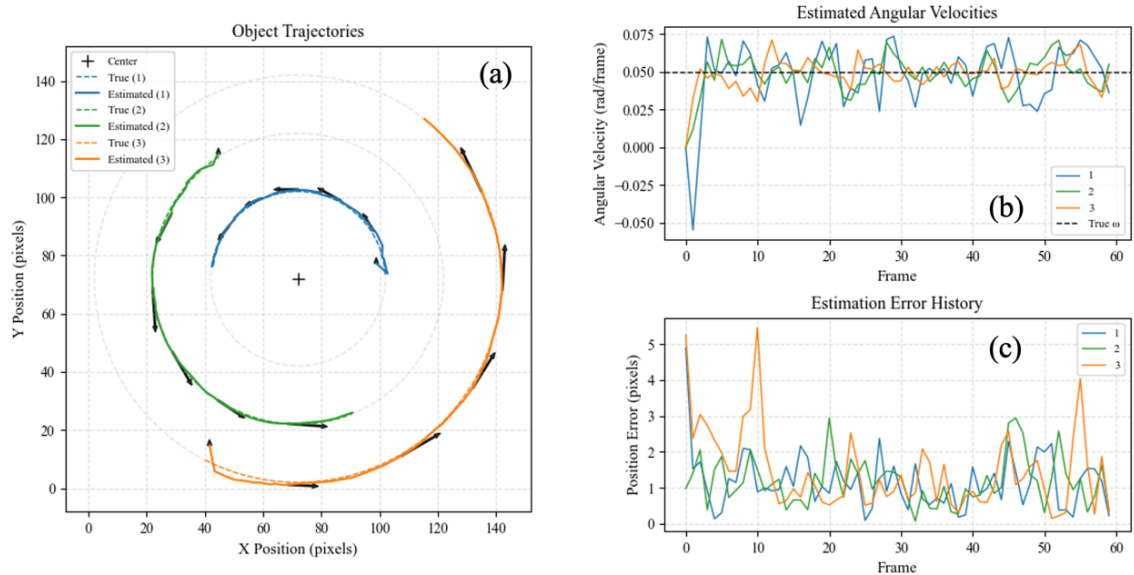

**Fig. 3  Tracking performance of the proposed framework. (a) True and estimated object trajectories. (b) Estimated angular velocities compared to ground truth. (c) Position estimation error over time for each object.**

## IV.  Conclusion

This work presents a novel hybrid neuromorphic framework that fuses the strengths of Artificial Neural Networks (ANNs) and Spiking Neural Networks (SNNs) to enable real-time, energy-efficient obstacle detection and localization in robotic navigation systems. By adopting a dual-pathway design, the proposed system successfully separates static and dynamic sensory processing, assigning low-frequency spatial analysis to ANNs and high-frequency, event-driven temporal analysis to SNNs. This decoupling not only enhances the system's responsiveness to unmodeled and dynamic intruders but also minimizes the computational burden through the use of a pre-developed SNN-based filter, SNN-EMSIF, for robust state estimation. Simulation results validate the feasibility of this architecture, demonstrating promising accuracy in object localization and velocity estimation while retaining low-latency response and efficient computational performance. The framework's capability to operate without online learning and its robustness to partial observability and neuron failure make it particularly suitable for deployment in edge robotics and autonomous systems operating under dynamic and unpredictable environmental conditions.

Future research may explore the utilization of the proposed framework results for the decision making and path planning in obstacle avoidance scenarios in the control systems.